\documentclass{article}

\PassOptionsToPackage{numbers, compress}{natbib}

\usepackage[preprint]{neurips_2024}

\usepackage[utf8]{inputenc} 
\usepackage[T1]{fontenc}    
\usepackage{hyperref}       
\usepackage{url}            
\usepackage{booktabs}       
\usepackage{amsfonts}       
\usepackage{nicefrac}       
\usepackage{microtype}      
\usepackage{xcolor}         

\usepackage{graphicx, caption} 
\usepackage{svg}
\usepackage{caption}
\usepackage{subcaption}
\usepackage{amsmath, amssymb, amsfonts, amsthm}

\usepackage[export]{adjustbox}[2011/08/13]

\title{Herd: Using multiple, smaller LLMs to match the performances of proprietary, large LLMs via an intelligent composer}

\author{Surya Narayanan Hari \\
Department of Biology and Biological Engineering \\
California Institute of Technology\\
 \texttt{shari@caltech.edu}\\
\AND
Rex Liu\\
California Institute of Technology\\
\texttt{rexliu@caltech.edu}\\
\AND
Matt Thomson\\
Department of Biology and Biological Engineering\\
Program in Computational and Neural Systems\\
California Institute of Technology\\
 \texttt{mthomson@caltech.edu}}

\begin{document}
\maketitle
\begin{abstract}

Currently, over a thousand LLMs exist that are multi-purpose and are capable of performing real world tasks, including Q\&A, text summarization, content generation, etc. However, accessibility, scale and reliability of free models prevents them from being widely deployed in everyday use cases. To address the first two issues of access and scale, organisations such as HuggingFace have created model repositories where users have uploaded model weights and quantized versions of models trained using different paradigms, as well as model cards describing their training process. While some models report performance on commonly used benchmarks, not all do, and interpreting the real world impact of trading off performance on a benchmark for model deployment cost, is unclear. Here, we show that a herd of open source models can match or exceed the performance of proprietary models via an intelligent router. We show that a Herd of open source models is able to match the accuracy of ChatGPT, despite being composed of models that are effectively 2.5x smaller. We show that in cases where GPT is not able to answer the query, Herd is able to identify a model that can, at least 40\% of the time.

\end{abstract}

\section{Introduction}

Large language models have found novel ways to increase the number of use cases, such as by expanding the number of parameters, combining existing models to augment a single models' functionality and quanitizing large models to fit on smaller devices \cite{brown2020language, openai2023gpt4, hu_lora_2021, xiao_smoothquant_2023, bai_towards_nodate,he2022masked, taori2023stanford, bommasani2021opportunities,brown2020language,dettmers2023qlora}. The rapid expansion of model availability has created a significant challenge in practice, where corporations want to expose performant  LLM endpoints for their users, and have to spend time evaluating models to find the best one that works for them in practice. To overcome this problem, engineers often resort to proprietary models without knowing if there are open-source models available at a comparable performance standard. 

\begin{figure}
    \centering
    \includegraphics[width = \columnwidth]{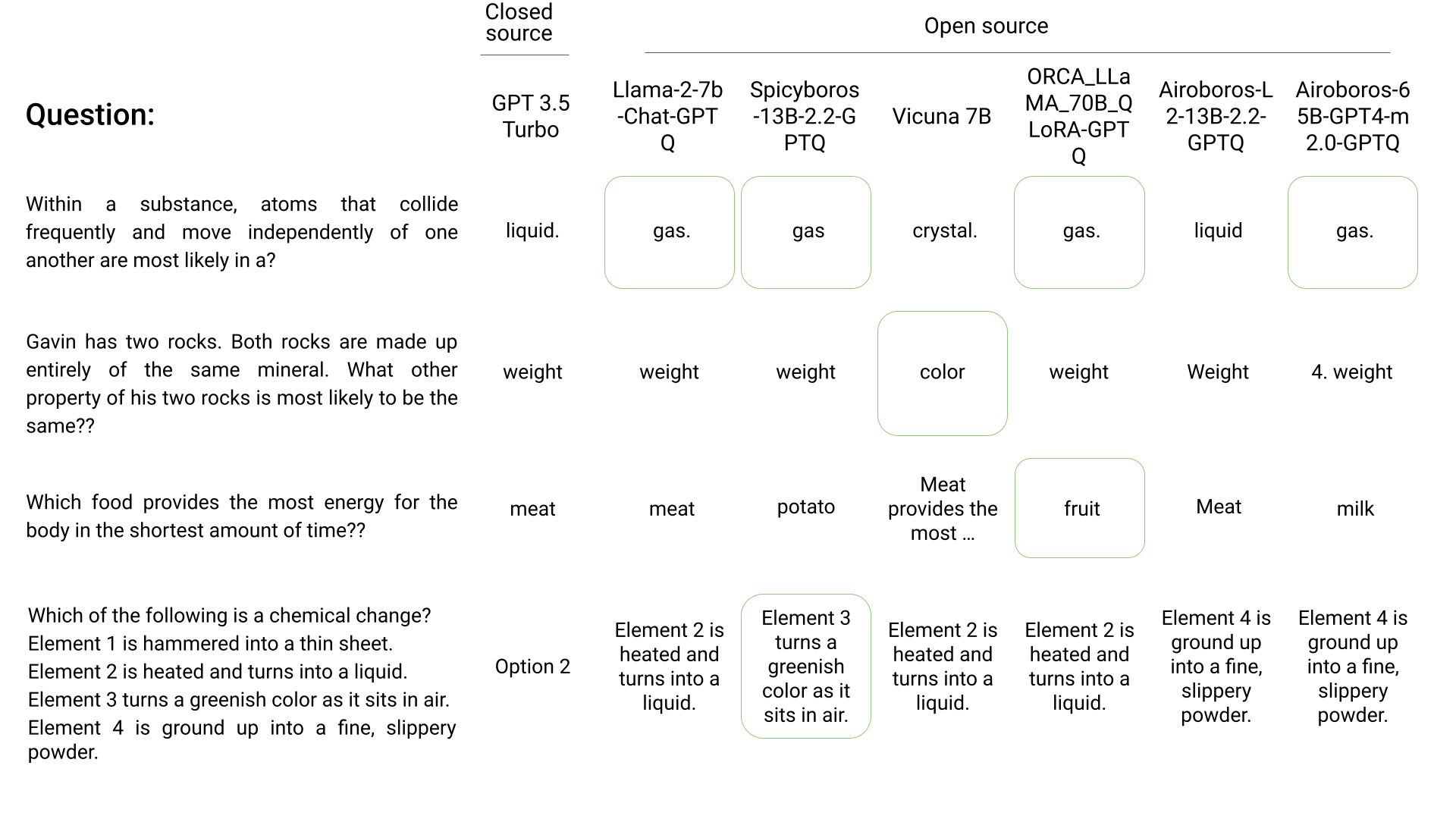}
    \caption{In practice, not all models are able to answer all questions accurately (the ones that do answer the questions correctly have their answers boxed in green), which leads to the practical challenge in picking an ensemble of models that has at least one highly performant model for every question. Herd attempts to solve this problem by constructing a herd of large language models that collectively can answer the query accurately, and by learning the association between input text and performance of each LLM.}
    \label{fig:Fig1}
\end{figure}

This often leads to the problem elaborated in Figure \ref{fig:Fig1}, showing examples of questions taken from MMLU that ChatGPT (GPT 3.5 Turbo) answers incorrectly, but there is some open source model that can answer the question correctly. We use this insight to try and construct a herd of models such that at least one model in the herd can answer any incoming query correctly. 

Recent model evaluation frameworks \cite{eval-harness, zheng2023judging} help users compare LLMs against each other, but the growing pace of model formats, outpaces one-size-fits-all comparison software suites. Empirical evidence in this work, reveals that open source models have caught up with leading proprietary models, but not all open source models feature on leaderboards, due to their vast number. 

Deployment of models also remains a key challenge. The 70b parameter Llama-2, in 16-bit precision, requires 2 80Gb A100 GPUs, and in practice, users might want several models running in parallel. Sacrificing parameter count to cut costs risks performance degradation, the exact magnitude of which is unknown before deployment. 

While quantized models might alleviate some of the challenges associated with model deployment, finding performant quantized models, navigating their formats and knowing their training details, such as what datasets were used in their quantisation calibration, requires expertise. 

In addition to quantized variants of models, specific model variants exist with chat capabilities, with different performance metrics from non-chat models. Others with more specific domain expertises such as science or code \cite{workshop_bloom_2023, noauthor_introducing_2023}, might be useful for some user applications but aren't fine-tuned for chat capability, making it harder to pick one model to use in production.

Today the Huggingface (HF) model repository contains $\sim$24,000 machine learning models for text generation. While model cards might provide some insight into the dataset that a model is trained on, common practices such as fine-tuning models using inputs from other large language models or model merging \cite{jin_dataless_2023,openchat,alpaca,mukherjee2023orca} has made it difficult to track what data was used to train the model. This has also made it challenging to track what datasets or tasks one can expect the models to be performant on. Futher, not all open source models have detailed model cards, making trusting them in deployment even more challenging.

Together, it would be a useful service to expose an endpoint that would process an incoming users' request by abstracting away model selection. Here, we explore the advantage of exposing a model herd of open source models, which outperforms a larger, proprietary large language model, offering size advantages. We also train a Tryage router \cite{hari_tryage_2023} to predict model performance, and show that the model herd is able to answer 74\% of incoming queries with performance comparable to or better than ChatGPT.

\section*{Herd Architecture}

Define a model Herd $M$, which is a collection of models. In an oracle model system, incoming query $z$ is assigned to $\underset{j}{\arg \max} \ M_j(z)$. However, in practice, evaluating $M_j(z)$ for all $j$, is expensive. To this end, we choose the model $\underset{j}{\arg \max} \ \hat{M}_j(z)$, where $\hat{M}_j$ is learned by a router $R$ \cite{hari_tryage_2023}. We implement the router model as a language model. In practice, the router optimizes its weight set $W$ over the following loss function. 
\begin{equation} \label{eq: predictive router}
\min_W \ \ \mathbb{E}_{z \sim p(z)} \ \left( \frac{1}{|M|} \sum_{M_i} D\big(R(z,M_i;W)||L(z,M_i)\big) \right)
\end{equation}

where $D(\cdot||\cdot)$ is divergence between predicted loss $R(z,M_i;W)$ and ground truth loss $L(z,M_i)$ (here the L1 distance function was used between predicted and ground truth F1s measured character-wise) for prompts $z$ drawn from data distribution $p(z)$ from the MMLU dataset. 

In this work, we found that bert-medium \cite{turc_well-read_2019} was the best performing router when a variety of router models were trained on 12,000 examples and validated on 3001 examples from MMLU, with a fixed batch size of 16, using the Adam optimizer with a learning rate of $2e-5$. 

In this work, we composed the herd by replicating realistic user constraints of models that would fit on an 8x48Gb cluster, using a mixture of 7B, 15B, 30B and 70B order models. We used a mix of quanitized and non-quantized models, since their performances' were previously unknown.

 
\section*{Demonstrating Herd}

\begin{figure}
    \centering
    \includegraphics[width = 0.5\textwidth]{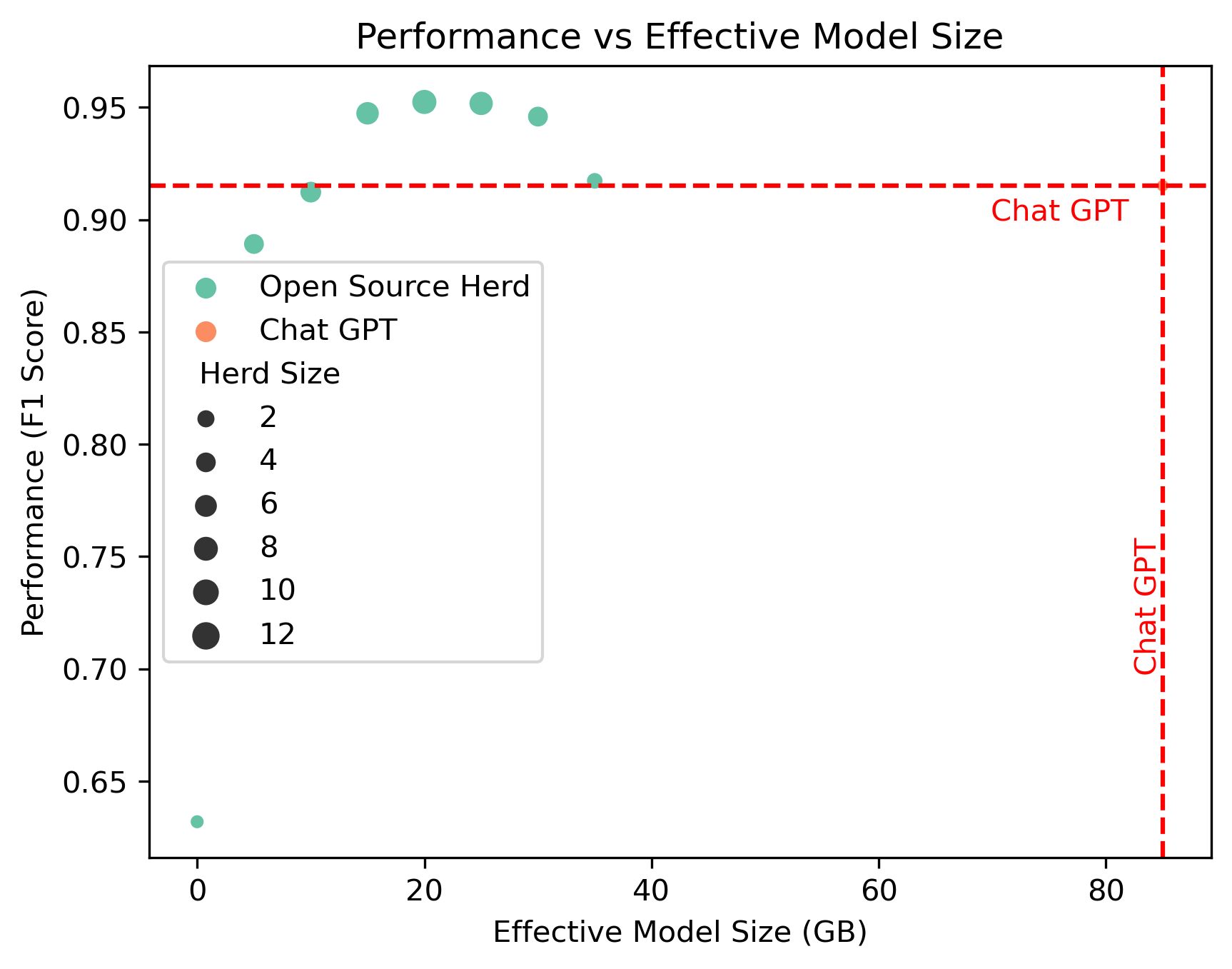}
    \caption{Open source model Herds outperform proprietary models such as ChatGPT on MMLU with decreased model size.}
    \label{fig:Fig2}
\end{figure}

We find that a herd of open source models is able to beat ChatGPT (Figure \ref{fig:Fig2}) despite being effectively less than 30\% of the size (effective size measured as the average size of models weighted by the number of examples allocated to them. Further, none of the models in the herd were individually better than ChatGPT, but together, they were able to surpass ChatGPT's performance. Further, all the models are open source, and the herd can be seamlessly expanded, contracted or interchanged for other models. 

\begin{figure}
     \centering
     \begin{subfigure}[b]{0.45\textwidth}
         \centering
         \includegraphics[width=\textwidth]{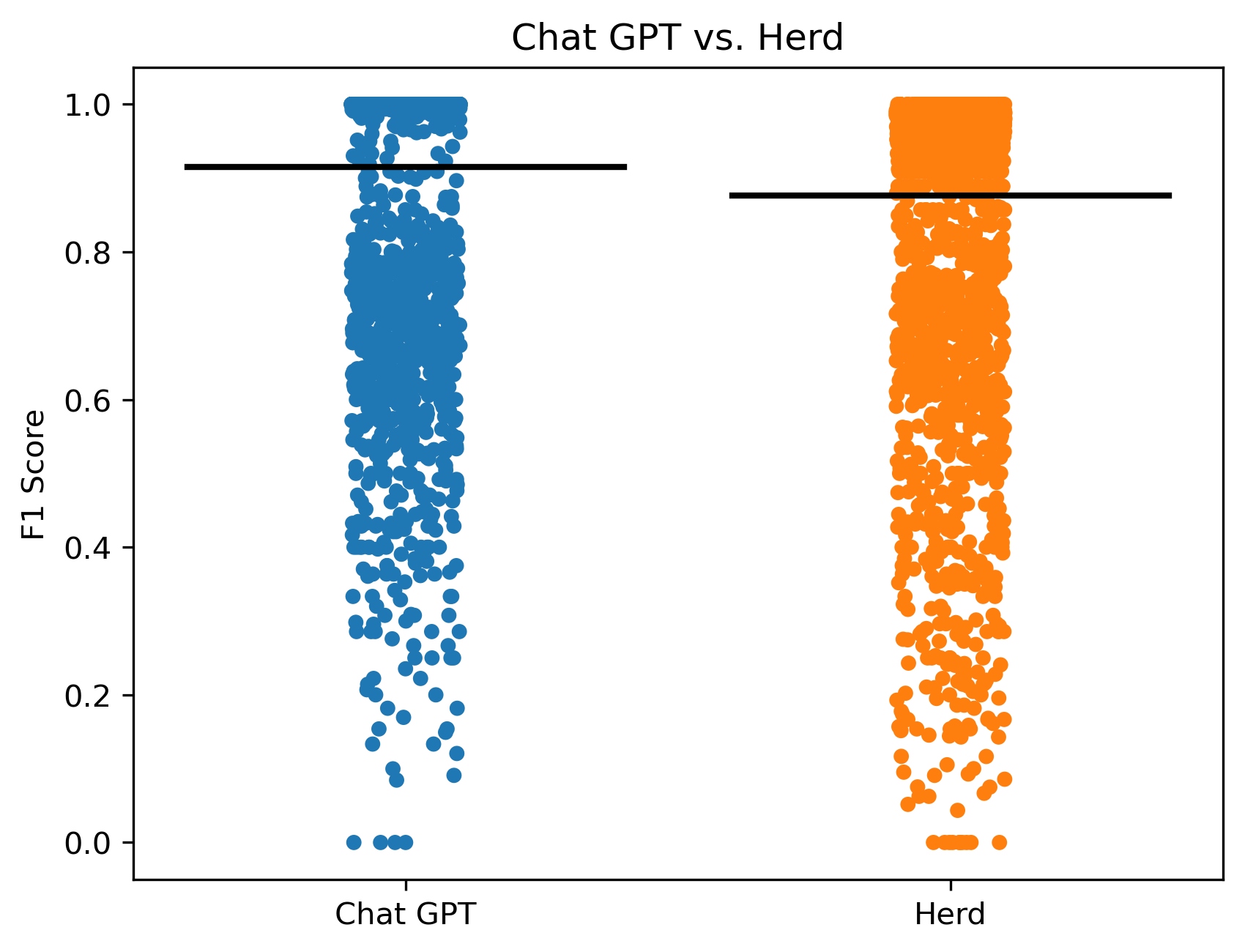}
            \caption{}
        \label{fig:3a}
     \end{subfigure}
     \hfill
     \begin{subfigure}[b]{0.45\textwidth}
         \centering
         \includegraphics[width=\textwidth]{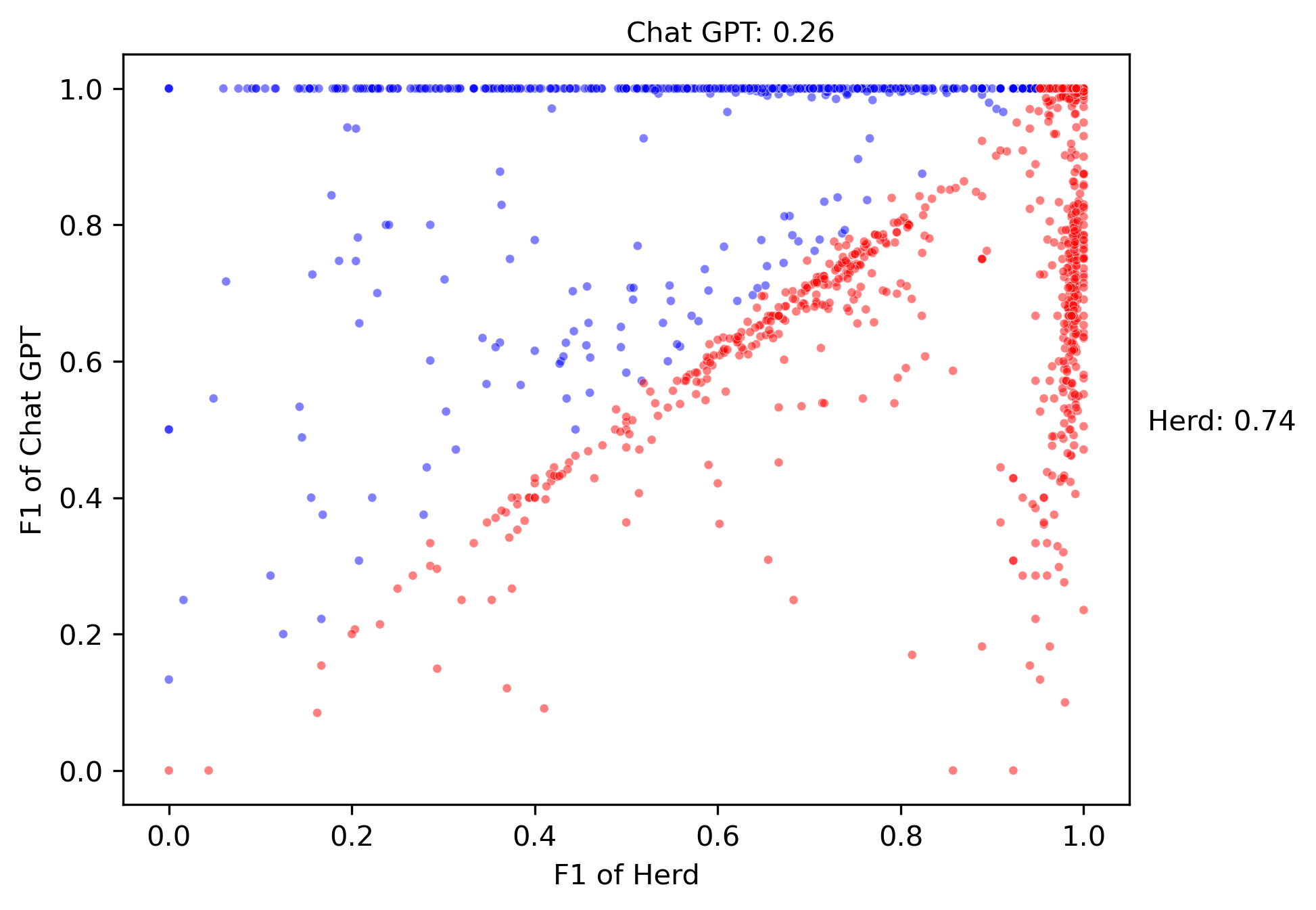}
         \caption{}
         \label{fig:3b}
     \end{subfigure}
     \hfill
     \begin{subfigure}[b]{0.8\textwidth}
         \centering
         \includegraphics[width=\textwidth]{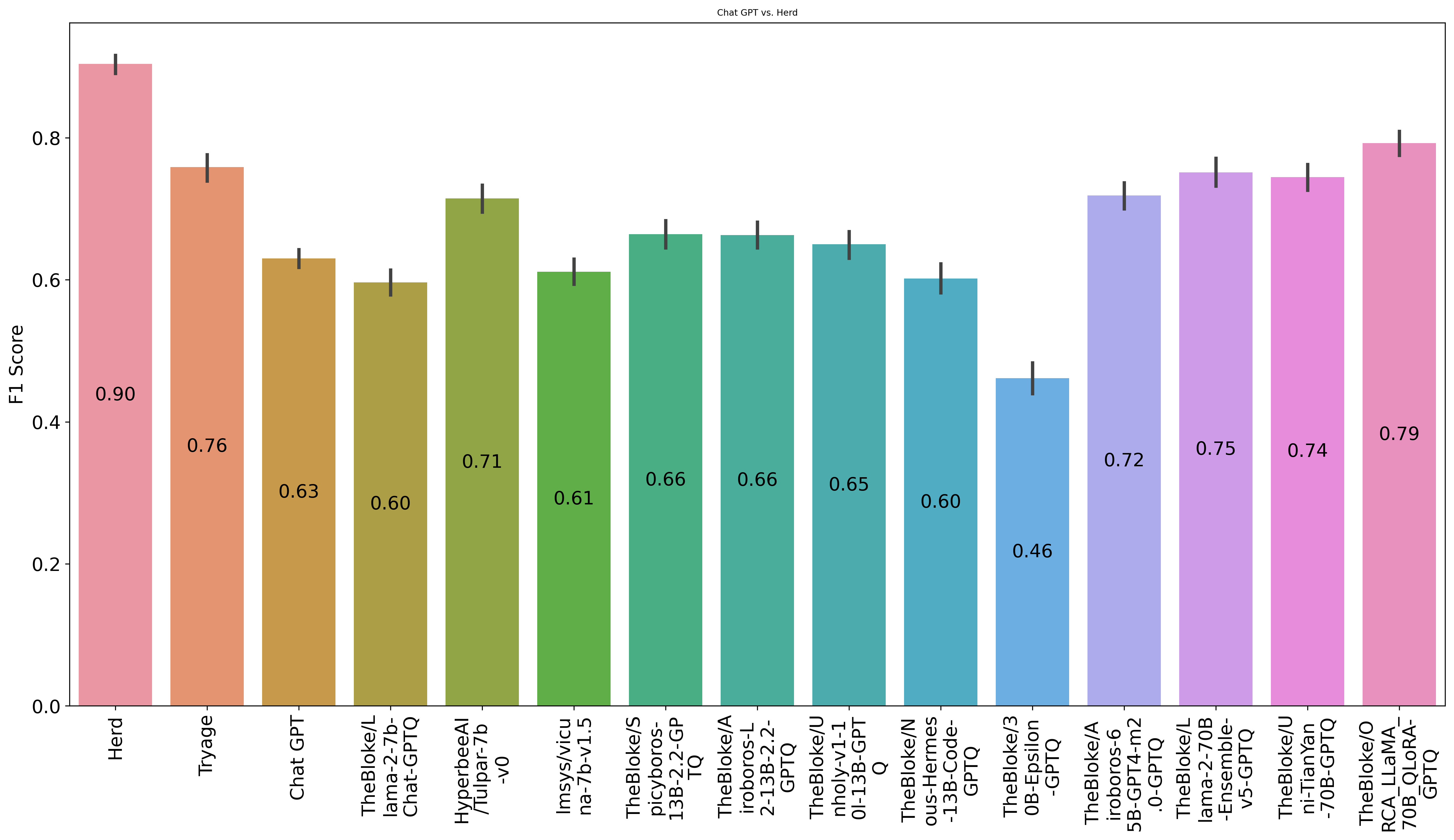}
         \caption{}
         \label{fig:3c}
     \end{subfigure}
        \caption{a) A router trained to model the performance of a herd offers comparable performance to GPT 3.5 Turbo (mean performances shown as horizontal lines). b) GPT exceeds the performance of the Herd in only 26\% of incoming queries, implying 74\% of incoming queries can be answered by open source models in the Herd. c) In questions that ChatGPT gets wrong the Herd can find models that perform correctly (Average of 0.9 F1). A routing model, achieves an aggregate of 0.76 F1 on these questions. }
        \label{fig:Fig3}
\end{figure}

We trained a tryage router \cite{hari_tryage_2023} to model the performances of a herd and found that the router was able to successfully allocate incoming queries to models that produced  aggregate performance comparable to GPT 3.5 Turbo despite being effectively 2.5x smaller \ref{fig:3a} \footnote{exact number of parameters in ChatGPT (GPT 3.5 Turbo) unknown, based on reported information}. Further, some models in the herd are quantized, meaning they can be run on edge compute / cloud compute - a user can trade off the size of a herd for compute cost. 

We show that Herd can capture knowledge in cases where ChatGPT fails to answer an incoming query. While any single model might not be able to answer all the incoming queries, Herd is able to find a model that can answer each query, based on the input text of the prompt. ChatGPT is only able to beat a herd of open source models 26\% of the time, implying 74\% of the queries can be answered by open source models (Fig. \ref{fig:3b}, `beat' is defined as F1 in excess of 5\%). 

In the cases where ChatGPT was wrong, defined as when ChatGPT had an F1 score of less than 0.9, Herd was able to achieve a correct answer (defined as when any model in the Herd had an F1 score greater than 0.95), 69.3\% of the time. A predictive router, was able to identify a model that can answer the query correctly, 40\% of the time (Tryage bar in Fig. \ref{fig:3c}). The mean of the F1s of the answers from each model, as well as the aggregate F1s from Herd and the predictive router, are shown in Figure \ref{fig:3c}.

\section*{Conclusion and discussion}
In this work we present the result that a Herd of open-sourced models can achieve performance comparable or better than ChatGPT, at a fraction of the compute cost and zero query cost. Further, when proprietary models cannot answer a query, a herd of open source models, are able to cover a significant portion of the deficit. This system offers a new model paradigm to compete against closed source models, by leveraging widely available open source technology.

\clearpage

\bibliographystyle{plain} 
\bibliography{References/references,References/scibib, References/surya-zotero, References/surya-refs-no-zotero}

\end{document}